\documentclass[conference]{IEEEtran}
\usepackage{multirow}

\ifCLASSINFOpdf
   \usepackage[pdftex]{graphicx}
\else
\fi
\begin{document}
%
\title{Optimisation of the PointPillars network for 3D object detection in point clouds}

\author{
	\IEEEauthorblockN{Joanna Stanisz}
	\IEEEauthorblockA{
		AGH University of Science \\
		and Technology Krakow, Poland \\
		E-mail: stanisz@student.agh.edu.pl}
	\and
	\IEEEauthorblockN{Konrad Lis}
	\IEEEauthorblockA{
		AGH University of Science \\
		and Technology Krakow, Poland \\
		E-mail: lis@student.agh.edu.pl}
	\and
	\IEEEauthorblockN{Tomasz Kryjak, \textit{Senior Member IEEE}}
	\IEEEauthorblockA{ AGH University of Science \\
		and Technology Krakow, Poland \\
		E-mail: tomasz.kryjak@agh.edu.pl}
	\and
	\IEEEauthorblockN{Marek Gorgon, \textit{Senior Member IEEE}}
	\IEEEauthorblockA{ AGH University of Science \\
		and Technology Krakow, Poland \\
		E-mail: mago@agh.edu.pl}		
}


%


\maketitle

\begin{abstract}

In this paper we present our research on the optimisation of a deep neural network for 3D object detection in a point cloud.
Techniques like quantisation and pruning available in the Brevitas and PyTorch tools were used.
We performed the experiments for the PointPillars network, which offers a reasonable compromise between detection accuracy and calculation complexity.
The aim of this work was to propose a variant of the network which we will ultimately implement in an FPGA device. 
This will allow for real-time LiDAR data processing with low energy consumption.
The obtained results indicate that even a significant quantisation from 32-bit floating point to 2-bit integer in the main part of the algorithm, results in 5\%-9\%  decrease of the detection accuracy, while allowing for almost a 16-fold  reduction in size of the model.


\end{abstract}


%
\IEEEpeerreviewmaketitle

\section{Introduction}

The detection of objects, i.e. vehicles, pedestrians, cyclists, animals etc. is crucial in advanced driver assistance systems (ADAS) and autonomous vehicles (AV).
It can be based on data from radars, cameras and LiDARs (Light Detection and Ranging).
The first two sensors are quite commonly used in vehicles currently (2020) available on the market and equipped with ADAS solutions.
However, fully autonomous vehicles (4 and 5th level of SAE classification) are often additionally equipped with the LiDAR sensor.
The most well-known examples are vehicles from Waymo \footnote {they are also the vehicles with the largest number of autonomously driven kilometres}, but most major car manufacturers and component manufacturers (Bosch, Aptiv) have test vehicles equipped with these sensors.
The advantages of the LiDAR, which is an active sensor, are low sensitivity to lighting conditions (including correct operation at nighttime) and a~fairly accurate 3D mapping of the environment, especially at a~short distance from the sensor.
The disadvantages include improper operation in the case of heavy rainfall or snowfall and fog (laser beam scattering occurs), deterioration of the image quality along with the increasing distance from the sensor (sparsity of the point cloud) and a~very high cost.
The last issue is the major reason that prevents this technology from being used more widely in commercially available vehicles.
However, it should be noted that there is constant progress in this area, including so-called solid-state solutions (without moving parts).
Therefore, one should expect that the LiDAR cost will be much lower soon.
The output from a~LiDAR sensor is a~point cloud, usually in the polar coordinate system.
A~reflection intensity coefficient is assigned to each point. 
Its value depends on the properties of the material from which the beam was reflected.

Because of the rather specific data format, object detection and recognition based on the LiDAR point cloud significantly differs from the methods known from ``standard'' vision systems.
Generally, two approaches can be distinguished: ``classic'' and based on deep neural networks.
In the first, the input point cloud is subjected to pre-processing (e.g. ground removal), grouping (using clustering or fixed three-dimensional cells), feature vector calculation and classification.
These methods achieve only moderate accuracy on widely recognised test data sets -- i.e. KITTI \cite {Kitti}, \cite {Kitti_www}.


In the second case, deep convolutional neural networks are used.
They provide excellent results (cf. the KITTI ranking \cite {Kitti_www}).
However, the price for the high accuracy is the computational and memory complexity, and the need for high-performance graphics cards (GPU) -- for training and inference.
This stands in contrast with the requirements for systems in autonomous vehicles, where the aim is to reduce the energy consumption while maintaining the real-time operation and high detection accuracy.


Recently, a~very promising research direction in embedded deep neural networks is the calculation precision reduction (quantisation). 
In many publications, it has been shown that the transition from a~32-bit or 64-bit floating point representation to a~fixed point and in an extreme cases even to a~binary one, results in a~relatively small loss of precision, and a~very significant reduction in computational and memory complexity.
For example, in \cite{Courbariaux_2016} experiments with a~binary and in \cite {Deng_2018} with ternary network were presented.
In addition, it is possible to use pruning (removing less important neurons), which allows to further reduce the computational complexity of the network.


In this work we evaluate the possibility of applying the above-mentioned optimisations to a~deep neural network for object detection on point cloud data from a~LiDAR sensor.
Based on the initial analysis, we selected the PointPillars \cite{pointpillars} network, mainly due to the favourable ratio of detection precision to the computational complexity.
Then, using the Brevitas \cite {brevitas} and PyTorch libraries, we conducted a~series of experiments to determine how limiting the precision and pruning affects the detection precision.
We were able to obtain almost a 16-fold reduction in the size of the model, by changing the precision from 32-bit floating-point to 2-bit integer.
This resulted in 5\%-9\%  decrease of the detection accuracy, which should be regarded as moderate and acceptable.
To our knowledge, similar experiments for this network have not been previously described in the literature.
This research is a~stage in the work on implementing the PointPillars network in an FPGA (Field Programmable Gate Array) device. 
This would allow to create an energy-efficient and working in real-time embedded object detection system.



The reminder of this paper is organised as follows.
In Section \ref{sec:prev_work} the general scheme of an object detection system based on data from the LiDAR sensor, the methods described in the literature, and the available LiDAR data sets are discussed.
Section \ref{sec:pointpillars} presents the used PointPillars network.
The applied optimisation methods, tools, comparison with similar approaches, and the obtained results are presented in Section \ref{sec:optimization}.
The paper ends with a~summary and an indication of further research directions.


\section{Object detection from a~LiDAR point cloud}
\label{sec:prev_work}

Object detection from a~LiDAR can be divided into several stages.
The first is data acquisition from the sensor followed by an optional coordinate system change --  from polar to Cartesian.
The next two stages are pre-processing: the detection and removal of the ground and data filtration (removal of single points).
An example ground removal algorithm is presented in \cite{Cabanes}.
The next step is data segmentation.
Its output are separate groups of points that will be classified.
The final stage is object detection and recognition.
In the classical approach it consists of: feature extraction, classification and final object grouping.
For the feature vector, the most commonly used quantities are: real cluster dimensions, the number of points, elements of the covariance matrix, elements of the inertia tensor, central moments or reflectance intensity histogram.
For classification the Support Vector Machine algorithm is often used \cite {Tang}.



In recent years, the detection and recognition stages have been often implemented using deep convolution neural networks.
Similar to image processing, these methods integrate virtually all processing steps, including feature extraction and classification.
They are characterised by high computational and memory complexity, but also assure high recognition performance.
We can make the following breakdown among neural networks for LiDAR data processing:
\begin{itemize}
    \item 2D methods -- the point cloud is projected onto one or more planes, which are then processed by typical convolutional networks -- e.g. the MV3D method \cite{MV3D},
    \item 3D methods -- the point cloud is processed without reducing the third dimension, the following subdivision can be made:
    \begin{itemize}
        \item methods operating on points -- these methods perform semantic segmentation or classify the entire cloud as an object -- an exemplary method is PointNets \cite {fpointnet},
        \item methods operating on cells -- these methods divide the three-dimensional space into cells (fixed size), aggregate the features of particular points into a~features vector for a~given cell and process the matrix of cells with 2D or 3D convolutional networks -- examples are VoxelNet \cite{voxelnet} and  PointPillars \cite{ pointpillars} (described in more detail in Section \ref{sec:pointpillars}),
        \item hybrid methods -- methods partly using both of the above described approaches -- an example is PV-RCNN  \cite{PVRCNN}.
    \end{itemize}
\end{itemize}


The main advantage of the PointPillars solution in comparison to other methods operating on cells is the use of 2D instead of 3D convolutions (like in VoxelNet \cite{voxelnet}).
This significantly reduces the computational complexity of the system, while maintaining the detection accuracy.
Therefore, we decided to start research on its acceleration.


In the scientific community working on object detection for autonomous vehicles the following databases are used: KITTI, Waymo Open Dataset and NuScenes.
The most popular of them is the KITTI Vision Benchmark Suite \cite {Kitti}, which was created in 2012.
Besides the point cloud from the LiDAR sensor, images from four cameras are also available in the data set: two monochrome and two colour ones, and information from the GPS/IMU navigation system.
In the object detection category, an important element of the database is the training data set containing $7481$  photos along with the corresponding point clouds and annotated objects.
In addition, KITTI keeps a~ranking of object detection methods in categories: BEV (bird eye view) and 3D.
In the first case, the output of the algorithm is compared to a~rectangle describing the object in the top view (3D data is projected into 2D).
In the second, the output is compared with a~cuboid describing the object in 3D.
In addition, the test cases are divided into three levels of difficulty.

In the experiments described in Section \ref{sec:optimization} we decided to use the KITTI data set due to the following reasons.
KITTI is still the most widely used LiDAR database.
This is because of a~ranking, which is highly recognised in the scientific community and contains results for many algorithms.
Thanks to this, it is easy to compare a~new solution with those proposed so far.
In addition, the PointPillars network was originally trained and evaluated on this set.
We plan to use the other data sets in our future research.


\section{The PointPillars network}
\label{sec:pointpillars}

\begin{figure*}[!t]
\centering
\includegraphics[width=1\textwidth]{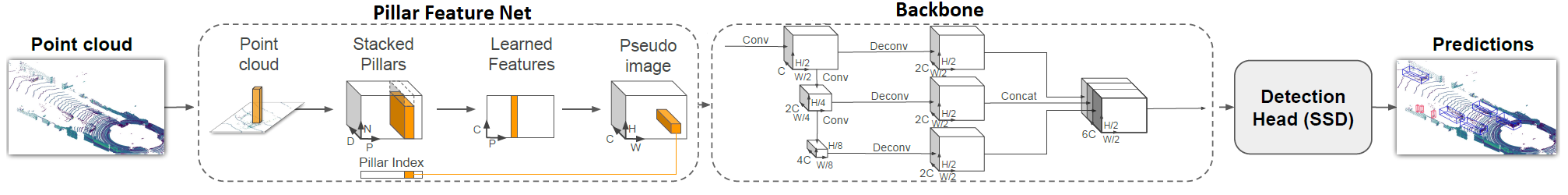}
\caption{An overview of the PointPillars networks structure \cite{pointpillars}.}
\label{fig:pointpillars}
\end{figure*}

The input to the PointPillars \cite{pointpillars} algorithm is a~point cloud from a~LiDAR sensor limited to the area located in front of the vehicle.
The results are oriented cuboids denoting the detected objects: cars, pedestrians and cyclists.
A~``pillar'' is a~three-dimensional cell, without a~user-defined height. 
It is created by dividing the point cloud in the XY plane (all operations are carried out in the Cartesian coordinate system).
An overview of the network structure is shown in Figure \ref{fig:pointpillars}.


The first part -- \textit{Pillar Feature Net} (PFN) -- converts the point cloud into a~sparse ``pseudo-image''.
Initially, the input data is divided into pillars.
The points in each of the pillars are then extended to a~nine-dimensional space ($D = 9$) by adding to the basic four point parameters (coordinates $x$, $y$, $z$ and reflection intensity) the values: $x_c$, $y_c$, $z_c$, $x_p$, $y_p$.
The variables with the $c$ designation describe the distance of the considered point to the centre of gravity of the points forming the pillar, while with the $p$ designation, the distance from the geometric centre of the pillar.
Because of the sparsity of the LiDAR data, most of the pillars do not contain any points.
For this reason, only a~few of nonempty pillars ($P$) form the input to the network. 
This approach reduces the memory complexity.
Additionally, a~limit on the number of points ($N$) in a~pillar is introduced to minimise the differences between very dense and sparse pillars.

The pillars are therefore fed to the network in the form of a~dense tensor with dimensions $(D,P,N)$.
Then, each point ($D$ dimensional) is processed by a~linear layer with \textit{batch normalization} and \textit{ReLU} activation function resulting in a~tensor with dimensions $(C,P,N)$.
Next, for each cell, all points are processed by a~\textit {max-pooling} layer creating a~$(C,P)$ output tensor.
Then it is mapped to a~$(C,H,W)$ tensor in such a~way, that the pillars are moved to their original location in the input cloud.
$H$ and $W$ are the dimensions of the pillar grid and simultaneously the dimensions of the ``pseudoimage''.

The second part of the network -- \textit{Backbone (2D CNN)} -- processes the ``pseudo-image'' and extracts high-level features.
It can be divided into two subnets: ``top-down'', which gradually reduces the dimension of the ``pseudoimage'' and another, which up-samples the intermediate feature maps and combines them into the final output map.
The ``top-down'' network can be described as a~series of blocks: \textit{Block $(S,L,F)$}.
A~block has $L$ convolution layers with a~3x3 kernel and $F$ output channels.
Each convolution is followed by a~\textit{batch normalisation} and a~\textit{ReLU} activation function.
The first layer in the block has a~$\frac{S}{S_{in}}$ step, while the next ones have a~step equal 1.
At the end of each block, the feature maps are up-sampled, from input stride $S_ {in}$ to output stride $S_ {out}$, using transposed convolution with $F$ output channels denoted as $ Up(S_ {in},S_{out}, F)$.
Then, after up-sampling, a~\textit {batch normalisation} and a~\textit{ReLU} activation function are used.
The final feature map is derived from the concatenation of all up-sampled output pillars feature maps.


The last part of the network is the \textit{Detection Head (SSD)}, whose task is to detect and regress the 3D cuboids surrounding the objects.
The objects are detected on a~2D grid using the Single-Shot Detector (SSD) network \cite{SSD} .
The position of the object along the $Z$ axis is derived from the regression map.
After inference, overlapping objects are merged using the Non-Maximum-Suppression (NMS) algorithm.


The detection results obtained using the PointPillars network in comparison with selected methods from the KITTI ranking are presented in Table  \ref{tab:bev_and_3d} (numbers taken from \cite{pointpillars}).
The list was limited only to car detection, because for these data the experiments described in Section \ref{sec:optimization} were carried out.
The AP (Average Precision) measure is used to compare the results:
$AP=\int_{0}^{1}p(r)dr$
where: $p(r)$ is the precision in the function of recall $r$.

\begin{table}[!t]
\caption{Comparison of the Average Precision results for the \textbf{BEV} and \textbf{3D} KITTI ranking (second column indicates the place in the ranking)}
\label{tab:bev_and_3d}
\begin{tabular}{c|l|l|l|l|l|l|}
\cline{2-7}
\multicolumn{1}{l|}{}                               & \multirow{2}{*}{\textbf{Place}} & \multirow{2}{*}{\textbf{Method}} & \multirow{2}{*}{\textbf{Data}} & \multicolumn{3}{c|}{\textbf{Car}}                \\ \cline{5-7} 
\multicolumn{1}{l|}{}                               &                                 &                                  &                                & \textbf{Easy}  & \textbf{Mod.}  & \textbf{Hard}  \\ \hline
\multicolumn{1}{|c|}{\multirow{4}{*}{\textbf{BEV}}} & 75                              & PointPillars                     & LiDAR                          & 88.35          & 86.10          & 79.83          \\ \cline{2-7} 
\multicolumn{1}{|c|}{}                              & 32                              & Patches                          & LiDAR                          & 89.72          & 89.39          & 83.19          \\ \cline{2-7} 
\multicolumn{1}{|c|}{}                              & 12                              & STD                              & LiDAR                          & 89.66          & 87.76          & \textbf{86.89} \\ \cline{2-7} 
\multicolumn{1}{|c|}{}                              & 2                               & PV-RCNN                          & LiDAR                          & \textbf{94.98} & \textbf{90.65} & 86.14          \\ \hline
\multicolumn{1}{|c|}{\multirow{4}{*}{\textbf{3D}}}  & 82                              & PointPillars                     & LiDAR                          & 79.05          & 74.99          & 68.30          \\ \cline{2-7} 
\multicolumn{1}{|c|}{}                              & 46                              & Patches                          & LiDAR                          & 88.67          & 77.20          & 71.82          \\ \cline{2-7} 
\multicolumn{1}{|c|}{}                              & 8                               & STD                              & LiDAR                          & 86.61          & 77.63          & 76.06          \\ \cline{2-7} 
\multicolumn{1}{|c|}{}                              & 1                               & PV-RCNN                          & LiDAR                          & \textbf{90.25} & \textbf{81.43} & \textbf{76.82} \\ \hline
\end{tabular}
\end{table}

When analysing the presented results, it is worth paying attention to the following issues.
First, progress in the field is rather significant and rapid -- the PointPillars method was published at the CVPR conference in 2019, and the PV-RCNN at CVPR in 2020.
Second, for the BEV case, the difference between these methods is about 7\%, and for the 3D case about 10\% -- this shows that the PointPillars algorithm does not very well regress the height of objects.
Third, the PV-RCNN network is much more complex than PointPillars.
Unfortunately, in \cite{PVRCNN}, the authors did not present data that would allow to describe this important parameter of the network unambiguously.

\section{Optimisation of the PointPillars network}
\label{sec:optimization}

\begin{table*}[!t]
\centering
\caption{Results of 3D detection with quantised Backbone and PFN layers.}
\label{tab:results1}
\begin{tabular}{|l|l|r|r|r|r|r|r|r|r|r|r|r|} 
\cline{3-13}
\multicolumn{1}{l}{}                &                     & \multicolumn{7}{c|}{\textbf{Backbone quantisation} }                                                                                                                                                                                            & \multicolumn{4}{c|}{\textbf{PFN quantisation} }                                                                                          \\ 
\hline
\multirow{4}{*}{\rotatebox{90}{\textbf{Network}}}  & \textbf{PFN}        & FP32                             & FP32                              & FP32                              & FP32                            & FP32                            & \textbf{FP32}                   & FP32                           & \textbf{INT16}  & \textbf{{INT8}}  & \textbf{INT4}  & \textbf{INT2}   \\ 
\cline{2-13}
                                    & \textbf{Backbone} & \textbf{FP32}  & \textbf{INT32}  & \textbf{INT16}  & \textbf{INT8}  & \textbf{INT4}  & \textbf{INT2}  & \textbf{BIN}  & INT2                              & INT2                            & INT2                            & INT2                             \\ 
\cline{2-13}
                                    & \textbf{SSD}        & FP32                             & FP32                              & FP32                              & FP32                            & FP32                            & \textbf{FP32}                   & FP32                           & FP32                              & FP32                            & FP32                            & FP32                             \\ 
\cline{2-13}
                                    & \textbf{Activation} & FP32                             & FP32                              & FP32                              & FP32                            & FP32                            & \textbf{FP32}                   & FP32                           & FP32                              & FP32                            & FP32                            & FP32                             \\ 
\hline
\multirow{3}{*}{\rotatebox{90}{\textbf{AP [\%]}}}   & \textbf{Easy}       & 80.19                            & 76.88                             & 76.34                             & 78.38                           & 74.24                           & \textbf{70.76}                  & 60.71                          & 72.72                             & 66.03                           & 62.38                           & 55.65                            \\ 
\cline{2-13}
                                    & \textbf{Moderate}   & 67.96                            & 66.38                             & 66.35                             & 67.45                           & 64.62                           & \textbf{61.60}                  & 47.09                          & 62.81                             & 54.03                           & 51.63                           & 44.03                            \\ 
\cline{2-13}
                                    & \textbf{Hard}       & 66.60                            & 60.28                             & 63.87                             & 65.48                           & 62.02                           & \textbf{55.94}                  & 44.96                          & 57.83                             & 53.29                           & 46.45                           & 43.22                            \\ 
\hline
\multirow{4}{*}{\rotatebox{90}{\textbf{Size[kiB]}}} & \textbf{PFN}        & 2.2                              & 2.2                               & 2.2                               & 2.2                             & 2.2                             & \textbf{2.2}                    & 2.2                            & 1.1                               & 0.6                             & 0.3                             & 0.1                              \\ 
\cline{2-13}
                                    & \textbf{Backbone}   & 18752.0                          & 18752.0                           & 9376.0                            & 4688.0                          & 2344.0                          & \textbf{1172.0 }                & 586.0                          & 1172.0                            & 1172.0                          & 1172.0                          & 1172.0                           \\ 
\cline{2-13}
                                    & \textbf{SSD}        & 30.1                             & 30.1                              & 30.1                              & 30.1                            & 30.1                            & \textbf{30.1}                   & 30.1                           & 30.1                              & 30.1                            & 30.1                            & 30.1                             \\ 
\cline{2-13}
                                    & \textbf{Sum}        & 18784.3                          & 18784.3                           & 9408.3                            & 4720.3                          & 2376.3                          & \textbf{1204.3}                 & 618.3                          & 1203.2                            & 1202.6                          & 1202.4                          & 1202.2                           \\
\hline
\end{tabular}
\end{table*}

\begin{figure}[!t]
\centering
\includegraphics[width=0.45\textwidth]{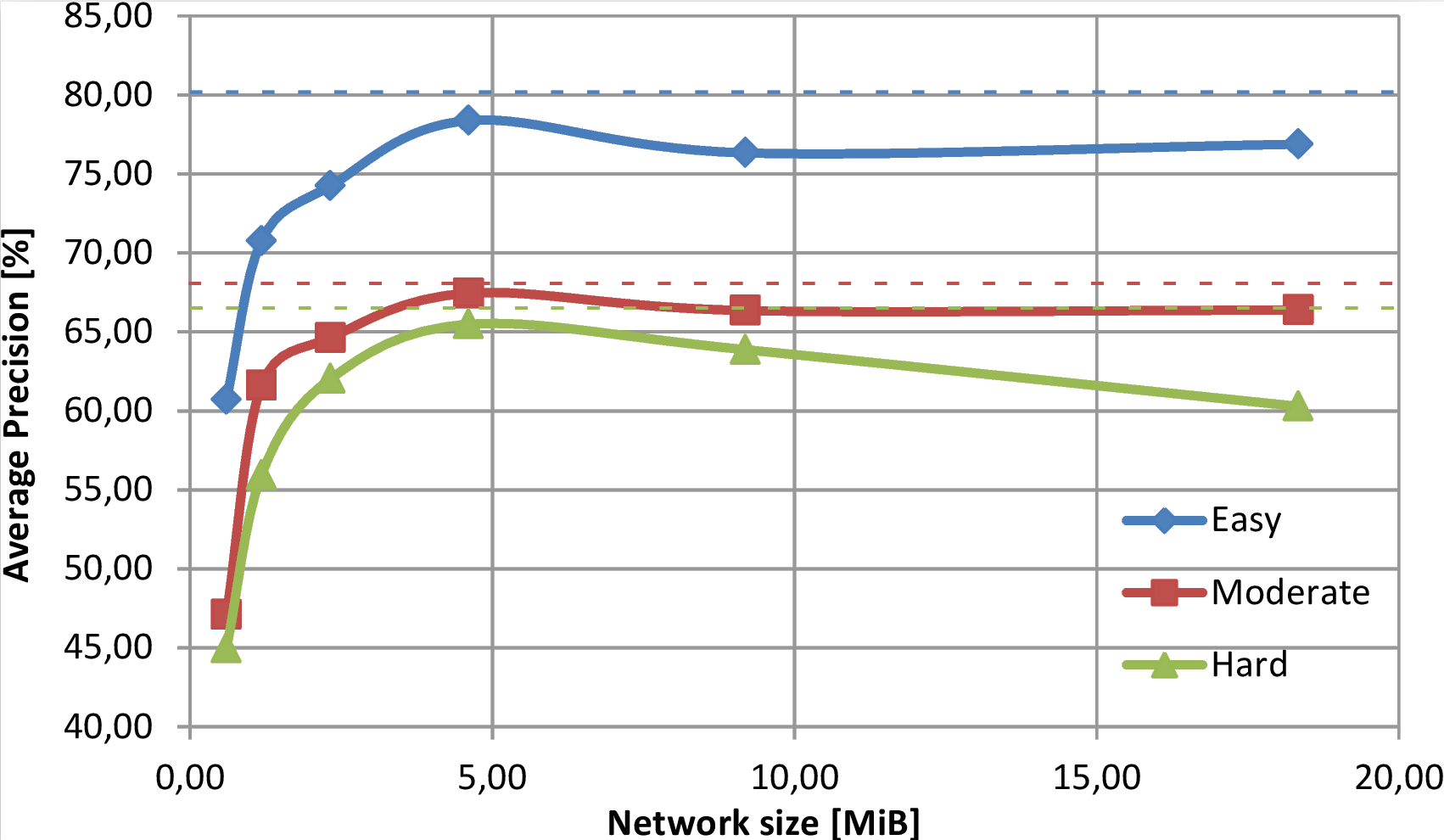}
\caption{Average Precision vs. network size}
\label{fig:results_AP_Size}
\end{figure}

\subsection{Optimisation methods}

There are basically three methods of optimising a~given deep network.
First, at the design stage, the number of layers can be reduced and simpler computational elements can be used.
This, however, contradicts the general rule that larger models, or models with additional modules, provide better results.
The PV-RCNN \cite{PVRCNN} network is a~good example of this issue.


Second, the required calculations can be quantized.
Initially, all arithmetic operations are carried out on single or double precision floating-point numbers (32 or 64-bit respectively).
This ensures high accuracy, but the parameters consume a~lot of memory resources and calculations require complicated hardware resources and thus much energy.
A~natural step, which has been applied for years in the hardware implementation of algorithms in reconfigurable devices (FPGA), is the transition to a~fixed-point representation with any precision from 64 bits to 1 bit only.
Both weights (network parameters) and activations (in this case ReLU function) can be quantised.
It is worth noting that quantisation is also used in dedicated hardware accelerators for artificial intelligence algorithms, e.g. Google's Tensor Processing Unit and many other accelerators.


The simplest solution, in the form of quantising the weights of a~trained model, does not provide good results -- a~significant decrease in the quality of operation is observed.
Therefore, training networks with limited precision is a~much better solution.
We used this approach in this paper.


The third popular method is the so-called pruning, i.e. removing connections and neurons with insignificant weights (i.e. with low value).
This idea was presented, for example, in the work \cite{lecun1990} from 1990.
In practice, this process consists of many iterations and requires multiple neuron removal and network re-training steps.


\subsection{The used tools}

As mentioned above, quantisation provides the best results when training networks from the beginning with quantised parameters.
In addition, it is important when implementing networks in embedded systems, including FPGA devices.
Therefore, in the research department of Xilinx, one of the leading manufacturers of reconfigurable devices, the FINN project was developed \cite{finn2020}, \cite{Blott_2018}.
It is an experimental framework that allows to analyse the possibilities of implementing quantized neural networks (QNN) in FPGAs.
It consists of three parts: the QNN network training tool (Brevitas), the FINN compiler that transforms the network description into a~hardware module that can be run in an~FPGA device and the Pynq-based deployment environment.



In the presented research we used the Brevitas tool.
It is a~library based on the popular PyTorch tool, which allows to train quantised neural networks (QNN).
It provides, among others, such options as choosing the type of quantisation (binary, ternary or integer), precision (constant or learned bit width) or scaling -- both for layers and the activation function.
The currently (2020) available layers include 1D and 2D convolutions and fully connected ones.
Brevitas does not define its own \textit{batch normalization} layer, but one can easily use the ones available in PyTorch.
The supported activation functions are: ReLU, Sigmoid, Tanh and HardTanh.


\subsection{PointPillars Quantisation}
\begin{table*}[!ht]
\centering
\caption{Results of 3D detection with quantised SSD and all layers.}
\label{tab:results2}
\begin{tabular}{|l|l|r|r|r|r|r|r|r|r|r|} 
\cline{3-11}
\multicolumn{1}{l}{}                &                                          & \multicolumn{4}{c|}{\textbf{SSD quantisation} }                                                                                         & \multicolumn{5}{c|}{\textbf{All layers quantisation} }                                                  \\ 
\hline
\multirow{4}{*}{ \rotatebox{90}{\textbf{Network}}} & \textbf{PFN}                             & \multicolumn{1}{l|}{FP32}         & \multicolumn{1}{l|}{FP32}       & \multicolumn{1}{l|}{FP32}       & \multicolumn{1}{l|}{FP32}       & \multicolumn{1}{l|}{INT16} & \multicolumn{1}{l|}{INT16} & \multicolumn{1}{l|}{INT16} & INT8   & INT8    \\ 
\cline{2-11}
                                    & \multicolumn{1}{r|}{\textbf{Backbone}} & INT2                              & INT2                            & INT2                            & INT2                            & INT2                       & INT2                       & INT2                       & INT2   & INT2    \\ 
\cline{2-11}
                                    & \textbf{SSD}                             & \textbf{INT16}  & \textbf{INT8}  & \textbf{INT4}  & \textbf{INT2}  & INT16                      & INT8                       & INT2   & INT8   & INT2    \\ 
\cline{2-11}
                                    & \textbf{Activation}                      & FP32                              & FP32                            & FP32                            & FP32                            & FP32                       & FP32                       & FP32                       & FP32   & FP32    \\ 
\hline
\multirow{3}{*}{\rotatebox{90}{\textbf{AP [\%]}}}   & \textbf{Easy}                            & 75.16                             & 75.69                           & 73.34                           & 73.01                           & 70.90                      & 70.75                      & 66.49  & 73.90  & 48.12   \\ 
\cline{2-11}
                                    & \textbf{Moderate}                        & 63.71                             & 65.26                           & 57.86                           & 62.99                           & 62.84                      & 62.66                      & 55.83                      & 63.42  & 40.60   \\ 
\cline{2-11}
                                    & \textbf{Hard}                            & 58.00                             & 59.04                           & 55.78                           & 57.14                           & 56.48                      & 56.45                      & 53.58  & 57.84  & 39.96   \\ 
\hline
\multirow{4}{*}{\rotatebox{90}{\textbf{Size[kiB]}}} & \textbf{PFN}                             & 2.2                               & 2.2                             & 2.2                             & 2.2                             & 1.1                        & 1.1                        & 1.1                        & 0.6    & 0.6     \\ 
\cline{2-11}
                                    & \textbf{Backbone}                        & 1172.0                            & 1172.0                          & 1172.0                          & 1172.0                          & 1172.0                     & 1172.0                     & 1172.0                     & 1172.0 & 1172.0  \\ 
\cline{2-11}
                                    & \textbf{SSD}                             & 15.0                              & 7.5                             & 3.8                             & 1.9                             & 15.0                       & 7.5                        & 1.9                        & 7.5    & 1.9     \\ 
\cline{2-11}
                                    & \textbf{Sum}                             & 1189.3                            & 1181.8                          & 1178.0                          & 1176.1                          & 1188.2                     & 1180.6                     & 1175.0                     & 1180.1 & 1174.4  \\
\hline
\end{tabular}
\end{table*}

To check how the PointPillars network optimisation affects detection precision (AP value) and the network size, we carried out several experiments.
We focused on data from the KITTI database, especially the car detection in the 3D category for three levels of difficulty: easy, medium and hard.
As a~reference we used a~network with all parameters in the 32-bit floating-point (FP32) representation.
We split the quantisation of the PointPillars network into four parts: Pillar Feature Net (PFN), Backbone, Detection Head (SSD), and activation function.


The original PointPillars network was trained using the Adam optimiser with initial learning rate $2*10^{-4}$ which decays by a~factor of 0.8 every 15 epochs.
The total number of epochs was originally 160, but during preliminary research we discovered that we can achieve satisfactory results after about 20.
Therefore, we decided to limit most experiments to 20 epochs — for such a~configuration, one experiment lasted an average of 3 hours on a~computer with AMD Ryzen 5 3600 processor and Nvidia RTX 2070S GPU.


We have divided the quantisation experiments into several stages.
The first one focused on the Backbone part because it has the largest impact on the size of the entire network (over 99\% for the base variant).
We summarise the results in the left part of the Table \ref{tab:results1} and in Figure \ref{fig:results_AP_Size}, where we present the AP value depending on the size of the network.
Dashed lines represent the AP reference value for the non-quantised network.
The waveforms for particular categories (easy, medium, hard) are similar - with a~network size smaller than 2 MiB (INT2 quantisation) there is a~significant decrease in accuracy, above 2 MiB the accuracy remains stable, with a~maximum of around 5 MiB (INT8 quantisation).
We have observed that the INT8 quantisation is closest to the original FP32, while the INT16 and INT32 quantisations resulted in noticeably worse results.
This observation requires further analysis, as it may result from the limited number of epochs used during training.


Further experiments focused on the INT2 quantisation of the Backbone network.
The network size was reduced by almost 16x, with an AP drop of max 11\%.
We have chosen this variant having in mind the future work on implementing this network in FPGA, where the size of the model is one of the key constraints.
In the case, when higher precision would be required, it is also worth considering the INT4 and INT8 variants, where the size reduction is 8x and 4x, respectively.


In the second step, we checked the effect of PFN quantisation on the AP parameter.
We summarise the results in the Table \ref{tab:results1}, columns \textit{PFN quantisation}.
Quantisation of this layer has a~big impact on the detection efficiency, and only negligible on the networks’ size.
For this reason, only the quantisation types INT16 and INT8 were considered in further experiments.


The third experiment examined how SSD quantisation affects AP.
We present the results in Table \ref{tab:results2}, \textit{SSD quantization} columns.
The layer is slightly larger than PFN, but also not significant from the point of view of the entire network (in relation to the INT2 variant it is only 2.5\%).
Interestingly, quantisation has a~relatively small impact on the final AP value, and we obtained the best results for the INT8 variant.


Based on the gathered results, we selected several quantisation options for the entire network and summarised them in Table \ref{tab:results2} in the \textit{All layers quantisation} columns.
We considered the variants INT16 and INT8 for the PFN, INT2 for the Backbone, and INT16, INT8 and INT2 for the SSD.
Then we re-trained the selected networks.
Ultimately, we obtained the best result for the PointPillars network in the variant $PP(INT8, INT2, INT8, FP32)$, where the individual elements represent the type of the PFN, the Backbone, the SSD and the activation functions quantisation, respectively.
Good results were also achieved for the configuration $PP(INT16, INT2, INT8, FP32)$.


For these two variants, we checked the effect of quantising the ReLU activation function.
We present the results of this experiment in Table \ref{tab:results3}.
We considered three variants INT16, INT8 and INT4.
It turned out that greater precision has a~negative impact on detection precision.
This issue requires further research.
Nevertheless, the differences between FP32 and INT4 activation for the considered network variants are small, and therefore we used this quantisation in further experiments.



Two of the considered variants were the most promising: $PP(INT8, INT2, INT8, INT4)$ and $PP(INT16, INT2, INT8, INT4)$.
We trained them for another 140 epochs to reach 160 in total.
The achieved results, including also $PP(FP32, FP32, FP32, FP32)$ trained for 160 epochs, are presented in Table \ref{tab:results_160epochs}.
The two considered variants have very similar AP results.
We finally selected $PP(INT8, INT2, INT8, INT4)$, as compared to the reference variant, it provides almost 16x lower memory consumption for weights while the AP value drops by max. 9\% in all three categories.




\begin{table}[t!]
\caption{Results of 3D detection for various network variants after training for 160 epochs}
\label{tab:results_160epochs}
\begin{tabular}{l|l|l|l|}
\cline{2-4}
                                                           & \textbf{Easy} & \textbf{Moderate} & \textbf{Hard} \\ \hline
\multicolumn{1}{|l|}{$PP(INT8, INT2, INT8, INT4)$}  & 76.64         & 67.33             & 65.56         \\ \hline
\multicolumn{1}{|l|}{$PP(INT16, INT2, INT8, INT4)$} & 76.71         & 66.45             & 65.20         \\ \hline
\multicolumn{1}{|l|}{$PP(FP32, FP32, FP32, FP32)$}  & 84.17         & 76.12             & 70.51         \\ \hline
\end{tabular}
\end{table}

\subsection{PointPillars pruning}



As part of further research, we considered pruning of the $PP(INT8, INT2, INT8, INT4)$ variant trained for 160 epochs.
The network was pruned in three variants -- 60\%, 70\% and 80\% of weights with the smallest absolute value from the entire network were zeroed, respectively (we used the functionality available in PyTorch).
Next we trained these variants with frozen and unfrozen pruned weights for another 20 epochs.
We show the results in Table \ref{tab:results_pruning}.
By saying frozen weights we mean keeping their values constant during the training (i.e. equal zero).
It can be seen that when zeroing 60\% and 70\% of the weights, the AP value remains the same.
When zeroing 80\% of the weights after pruning, we can see a~noticeable decrease in the AP value.
However, after re-training, the results are the best of the cases considered, moreover the unfrozen variants are comparable with the FP32 base variant trained for 20 epochs. 
Compared to the base variant trained for 160 epochs the AP drops by max. 8\% in all three categories.
The frozen variants have slightly decreased performance but still very close to $PP(INT8, INT2, INT8, INT4)$  trained for 160 epochs.
Using pruning does not directly affect the size of the network (``0’’ weights are also stored), but in the case of hardware implementation in an FPGA device, it allows for a~significant reduction of the necessary computational elements.


\begin{table}[!t]
\centering
\caption{Results of 3D detection with quantised activation functions.}
\label{tab:results3}
\begin{tabular}{|l|l|r|r|r|r|r|r|} 
\cline{3-8}
\multicolumn{1}{l}{}                 &                      & \multicolumn{6}{c|}{\textbf{Activation quantisation} }                                                 \\ 
\hline
\multirow{4}{*}{\rotatebox{90}{\textbf{Network}}} & \textbf{PFN}         & INT16           & INT16          & INT16          & INT8            & INT8           & INT8            \\ 
\cline{2-8}
                                     & \textbf{Back.}    & INT2            & INT2           & INT2           & INT2            & INT2           & INT2            \\ 
\cline{2-8}
                                     & \textbf{SSD}         & INT8            & INT8           & INT8           & INT8            & INT8           & INT8            \\ 
\cline{2-8}
                                     & \textbf{Act.}  & \textbf{INT16}  & \textbf{INT8}  & \textbf{INT4}  & \textbf{INT16}  & \textbf{INT8}  & \textbf{INT4}   \\ 
\hline
\multirow{3}{*}{\rotatebox{90}{\textbf{AP [\%]}}}    & \textbf{Easy}        & 63.10           & 69.28          & 72.60          & 57.47           & 72.53          & 72.05           \\ 
\cline{2-8}
                                     & \textbf{Mod.}    & 52.77           & 58.99          & 63.28          & 46.00           & 57.94          & 57.18           \\ 
\cline{2-8}
                                     & \textbf{Hard}        & 52.06           & 54.63          & 57.82          & 44.39           & 55.59          & 55.71           \\ 
\hline
\multirow{4}{*}{\rotatebox{90}{\textbf{Size[kiB]}}} & \textbf{PFN}         & 1.1             & 1.1            & 1.1            & 0.6             & 0.6            & 0.6             \\ 
\cline{2-8}
                                     & \textbf{Back.}    & 1172.0          & 1172.0         & 1172.0         & 1172.0          & 1172.0         & 1172.0          \\ 
\cline{2-8}
                                     & \textbf{SSD}         & 7.5             & 7.5            & 7.5            & 7.5             & 7.5            & 7.5             \\ 
\cline{2-8}
                                     & \textbf{Sum}         & 1180.6          & 1180.6         & 1180.6         & 1180.1          & 1180.1         & 1180.1          \\
\hline
\end{tabular}
\end{table}

\begin{table}[!]
\centering
\caption{Results of 3D detection after pruning with unfrozen and frozen weights.}
\label{tab:results_pruning}
\begin{tabular}{|l|l|l|l|l|l|} 
\cline{4-6}
\multicolumn{1}{l}{}                    & \multicolumn{1}{l}{}                        & \textbf{}      & \textbf{60\%}  & \textbf{70\%}  & \textbf{80\%}   \\ 
\hline
                                        & \multirow{3}{*}{\textbf{After pruning} }    & \textbf{Easy}  & 76.64          & 76.64          & 57.39           \\ 
\cline{3-6}
                                        &                                             & \textbf{Mod.}  & 67.33          & 67.33          & 50.92           \\ 
\cline{3-6}
\multicolumn{1}{|c|}{\textbf{Unfrozen}} &                                             & \textbf{Hard}  & 65.56          & 65.56          & 46.38           \\ 
\cline{2-6}
\textbf{weights}                        & \multirow{3}{*}{\textbf{After retraining} } & \textbf{Easy}  & 78.67          & 79.66          & 80.79           \\ 
\cline{3-6}
                                        &                                             & \textbf{Mod.}  & 67.30          & 67.41          & 68.68           \\ 
\cline{3-6}
                                        &                                             & \textbf{Hard}  & 65.56          & 65.66          & 66.60           \\ 
\hline
                                        & \multirow{3}{*}{\textbf{After pruning} }    & \textbf{Easy}  & 76.64          & 76.64          & 57.39           \\ 
\cline{3-6}
                                        &                                             & \textbf{Mod.}  & 67.33          & 67.33          & 50.92           \\ 
\cline{3-6}
\textbf{Frozen}                         &                                             & \textbf{Hard}  & 65.56          & 65.56          & 46.38           \\ 
\cline{2-6}
\textbf{weights}                        & \multirow{3}{*}{\textbf{After retraining} } & \textbf{Easy}  & 78.83          & 78.72          & 76.94           \\ 
\cline{3-6}
                                        &                                             & \textbf{Mod.}  & 68.71          & 68.16          & 66.84           \\ 
\cline{3-6}
                                        &                                             & \textbf{Hard}  & 66.78          & 66.12          & 64.94           \\
\hline
\end{tabular}
\end{table}

\section{Conclusion}
\label{sec:conclusion}

The article presents research on optimisation of the PointPillars network for 3D object detection based on a~LiDAR point cloud.
We used two techniques: quantisation and pruning.
We performed several experiments with different variants of network quantisation, with the PFN, the Backbone, the SSD and the activation function parts separately.
It turned out that the most important is the quantisation of the Backbone part, which is responsible for 99\% of the size of the input network.
In this case, changing the calculation precision from FP32 to INT2 resulted in almost a~16-fold reduction in size at a~cost of approx. 11\% AP (average precision) loss.
The quantisation of the other layers had a~small impact on the network size and ambiguous on the AP value.
We trained the best variant  $PP(INT8, INT2, INT8, INT4)$ for 160 epochs, which allowed to improve the precision by approx. 4\% for easy and 10\% for medium and hard test cases.
In addition, we carried out several pruning experiments which showed that the removal of 80\% of neurons followed by re-training results only in slight decrease of performance.
The final variant of the optimised PointPillars network, compared to the input one with floating-point precision, has the AP value decreased by max. 9\%, and after pruning with frozen weights by max. 10\%.
Most operations use INT2 numbers, some INT8, and the whole model is about 16 times smaller than the original one.
Summing up, the obtained results show that the optimisations used for PointPillars networks provide very good results, and an energy-efficient, real-time implementation in reprogrammable devices should be fully possible.

In the first stage of further work on the system we will use the FINN tool to generate a~hardware module with the optimised variant of the PointPillars network and try to run it on an FPGA or Zynq SoC device.
Ultimately, we would like to create a~demonstrator cooperating with a~LiDAR sensor.
Then, we plan to conduct experiments with the network for the other categories from the KITTI set, i.e. pedestrians and cyclists, as well as the Waymo and NuScenes sets.
In addition, we will carry out an in-depth quantisation analysis of the Backbone layer.
In the longer term, we consider modifying the network’s architecture using a~state-of-the-are research results or adding additional elements (like in \cite{PVRCNN}).
Ultimately, we would like to use data fusion for LiDAR, video and radar sensors.


\section*{Acknowledgment}
The work presented in this paper was supported by AGH University of Science and Technology project no. 16.16.120.773.

\end{document}